\documentclass[final]{cvpr}

\usepackage{times}
\usepackage{epsfig}
\usepackage{graphicx}
\usepackage{amsmath}
\usepackage{amssymb}

\usepackage{mmstyles}
\usepackage{subfigure}
\usepackage{multirow}
\usepackage{makecell}
\newtheorem{theorem}{Theorem}

\usepackage[pagebackref=true,breaklinks=true,colorlinks,bookmarks=false]{hyperref}

\def\cvprPaperID{3659} 
\def\confYear{CVPR 2021}

\begin{document}

\title{Towards Evaluating and Training Verifiably Robust Neural Networks}

\author{
Zhaoyang Lyu$^{1,2}$ \; Minghao Guo$^{1,2}$ \; Tong Wu$^{1,2}$ \; Guodong Xu$^{1,2}$ \; Kehuan Zhang$^{2}$ \; Dahua Lin$^{1,2,3}$\\
$^1$SenseTime-CUHK Joint Lab \; $^2$The Chinese University of Hong Kong\\
$^3$Centre of Perceptual and Interactive Intelligence\\
{\tt\small lyuzhaoyang@link.cuhk.edu.hk,
\{gm019, wt020, xg018, khzhang, dhlin\}@ie.cuhk.edu.hk}
}

\maketitle

\begin{abstract}
\thispagestyle{empty}
\vspace{-1em}
Recent works have shown that interval bound propagation (IBP) can be used to train verifiably robust neural networks. 
Researchers observe an intriguing phenomenon on these IBP trained networks: CROWN, 
a bounding method based on tight linear relaxation, often gives very loose bounds on these networks. 
We also observe that most neurons become dead during the IBP training process, which could hurt the representation capability of the network.
In this paper, we study the relationship between IBP and CROWN, and prove that CROWN is always tighter than IBP when choosing appropriate bounding lines. 
We further propose a relaxed version of CROWN, linear bound propagation (LBP), that can be used to verify large networks to obtain lower verified errors than IBP. 
We also design a new activation function, parameterized ramp function (ParamRamp), which has more diversity of neuron status than ReLU. We conduct extensive experiments on MNIST, CIFAR-10 and Tiny-ImageNet with ParamRamp activation and achieve state-of-the-art verified robustness. Code is available at \url{https://github.com/ZhaoyangLyu/VerifiablyRobustNN}.
\end{abstract}

\vspace{-1.5em}
\section{Introduction}
\vspace{-0.5em}
Deep neural networks achieve state-of-the-art performance in many tasks, \eg, image classification, object detection, and instance segmentation, but they are vulnerable to adversarial attacks. A small perturbation that is imperceptible to humans can mislead a neural network's prediction~\cite{szegedy2013intriguing, carlini2017towards,athalye2018obfuscated, kurakin2016adversarial, carlini2017adversarial}.
To mitigate this problem, Madry \etal~\cite{madry2018towards} develop an effective framework to train robust neural networks. They formulate adversarial training as a robust optimization problem. 
Specifically, they use projected gradient descent (PGD) to find the worst-case adversarial example near the original image and then minimize the loss at this point during training.
Networks trained under this framework achieve state-of-the-art robustness under many attacks~\cite{zhang2019theoretically, Wang2020Improving, rice2020overfitting}. 
However, these networks are only emperically robust, but not verifiably robust. They become vulnerable when stronger attacks are presented~\cite{wang2018mixtrain,croce2020reliable,tjeng2019evaluating}.
\begin{figure}[t]
   \centering
   \subfigure[Constant]{
    \label{fig:constant_bdl}
    \includegraphics[width=0.48\columnwidth]{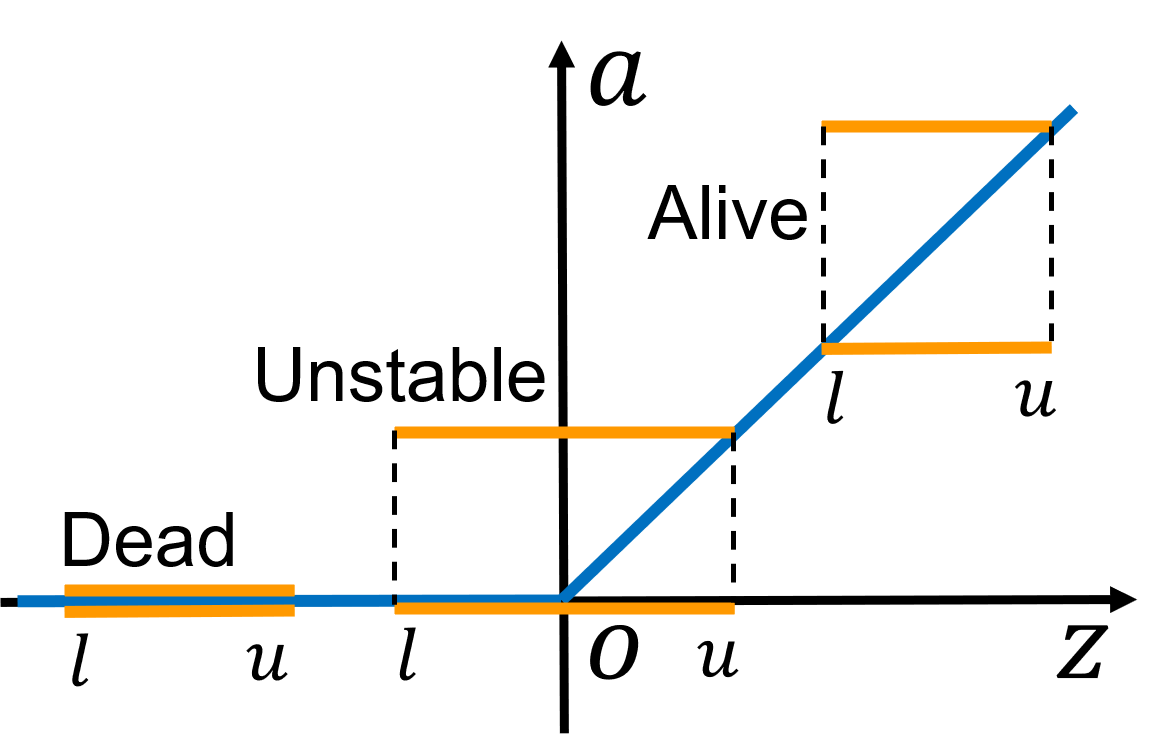}}
    \subfigure[Tight]{
    \label{fig:zero_bdl}
    \includegraphics[width=0.48\columnwidth]{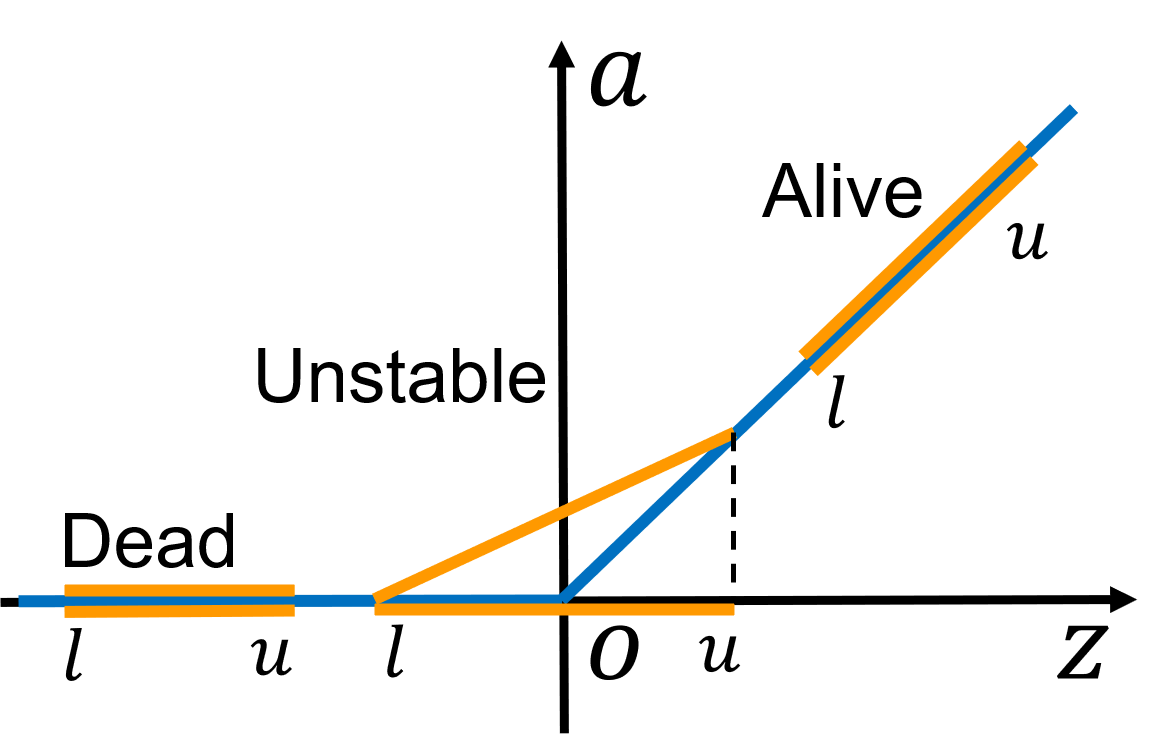}}
    \\
    \subfigure[Adaptive: Case $|l|>u$]{
    \label{fig:adaptive_bdl_1}
    \includegraphics[width=0.48\columnwidth]{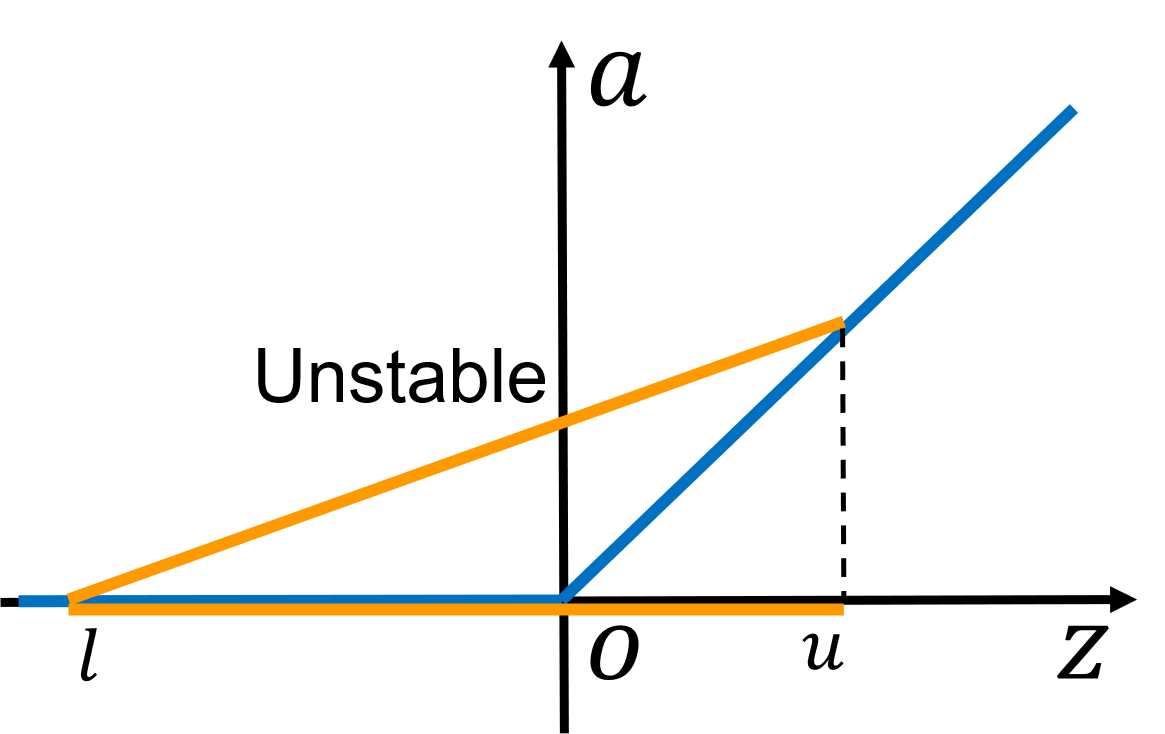}}
    \subfigure[Adaptive: Case $|l|\leq u$]{
    \label{fig:adaptive_bdl_2}
    \includegraphics[width=0.48\columnwidth]{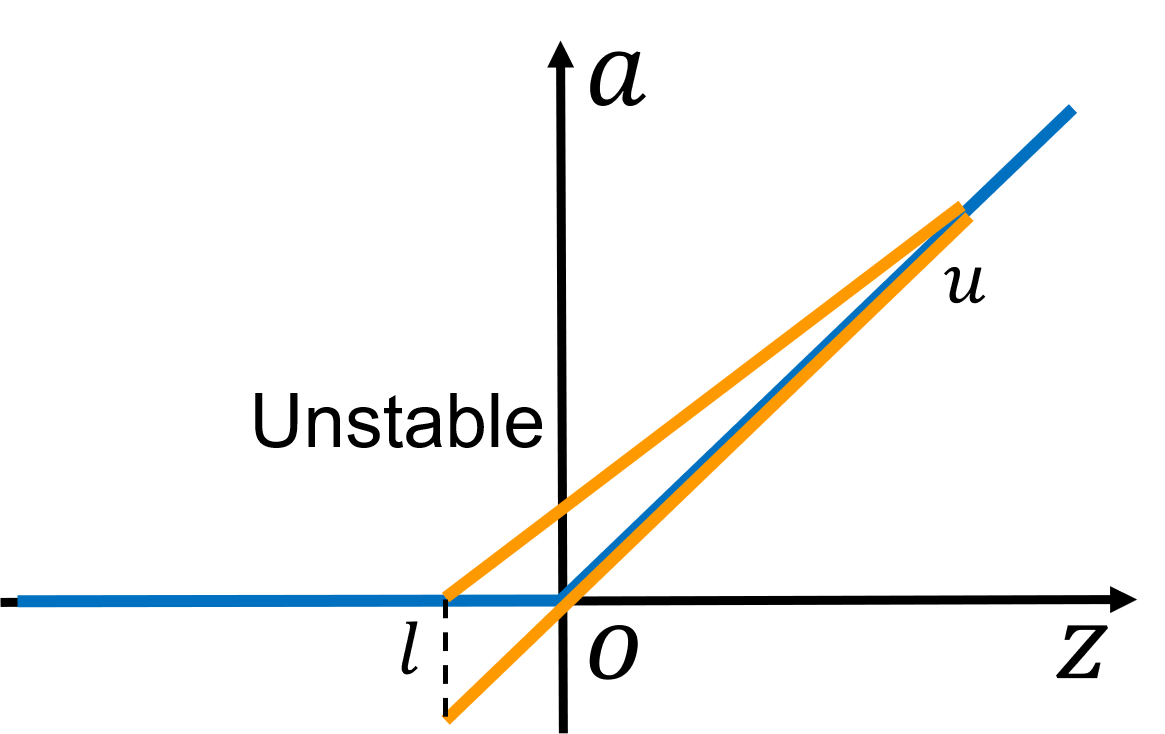}}
    \caption{Illustration of different strategies to choose bounding lines for the three status of a ReLU neuron. Dead: $l \leq u \leq 0$; Unstable: $l < 0 < u$; Alive: $0 \leq l \leq u$. $[l,u]$ is the input range of the neuron. 
    (a) chooses constant bounding lines.
    (b) is the tight strategy. 
    (c) and (d) are the two cases of unstable neurons in the adaptive strategy. The adaptive strategy chooses the same bounding lines as the tight strategy for dead and alive neurons. See more details in 
    {Appendix A.7}.}
    \vspace{-2em}
    \label{fig:bounding_line_strategies}
\end{figure}

This leads to the development of robustness verification, which aims to provide a certificate that a neural network gives consistent predictions for all inputs in some set, usually an $l_p$ ball around a clean image.
The key of robustness verification is to compute the lower and upper bounds of the output logits when input can take any value in the $l_p$ ball.
The exact bounds can be computed through Satisfiability Modulo Theory~\cite{katz2017reluplex} or solving a
Mixed Integer Linear Programming (MILP) problem~\cite{tjeng2019evaluating, cheng2017maximum}.
Relaxed bounds can be obtained by reduce the bound computation problem to a linear programming (LP) problem~\cite{wong2018provable} or a semidefinite programming (SDP) problem~\cite{dathathri2020enabling}.
However, these programming based methods are expensive and difficult to scale to large networks.
To this end, another approach that makes linear relaxations of the nonlinear activation functions in a network is proposed~\cite{singh2018fast, singh2019abstract, wang2018efficient, weng2018towards, zhang2018crown, ko2019popqorn}.
Figure~\ref{fig:bounding_line_strategies} illustrates different strategies to make linear relaxations of a ReLU neuron. 
These methods can compute bounds analytically and efficiently. 
In this paper, we focus on the study of CROWN~\cite{zhang2018crown}, which can \emph{compute relatively tight bounds while being fast}. Other similar approaches \cite{singh2018fast, singh2019abstract, wang2018efficient, weng2018towards} are either a special case of CROWN or a different view of it as demonstrated by Salman \etal~\cite{salman2019convex}.

Wong \etal~\cite{wong2018provable} propose to incorporate bounds computed by the aforementioned linear relaxation based methods in the loss function to train verifiably robust networks. Similar approaches are proposed in several other works~\cite{mirman2018differentiable, dvijotham2018training, raghunathan2018certified, wang2018mixtrain}. However, these methods generally bring heavy computational overhead to the original training process.
Gowal \etal~\cite{gowal2019scalable} propose to use a simple technique, interval bound propagation (IBP), to compute bounds. IBP is fast and can scale to large networks. Despite being loose, IBP outperforms previous linear relaxation based methods in terms of training verifiably robust networks. Zhang \etal~\cite{zhang2020towards} further improve this method by combining IBP with the tighter linear relaxation based method, CROWN. The resulting method is named CROWN-IBP. They use CROWN-IBP to compute bounds at the initial training phase and 
achieve the lowest $l_{\infty}$ verified  errors.

We notice that both IBP trained networks~\cite{gowal2019scalable} and CROWN-IBP trained networks~\cite{zhang2020towards} are verified by IBP after training. One natural question is whether we can use tighter linear relaxation based methods to verify the networks to achieve lower verified error. Surprisingly, Zhang \etal~\cite{zhang2020towards} find the typically much tighter method, CROWN, gives very loose bounds for IBP trained networks. It seems that IBP trained networks have very different verification properties from normally trained networks. 
We also find that CROWN cannot verify large networks due to its high memory cost.
Another phenomenon we observe on IBP and CROWN-IBP trained networks is that most neurons become dead during training. We believe that this could restrict the representation capability of the network and thus hurt its performance. In this paper, we make the following contributions to tackle the aforementioned problems:
\begin{enumerate}
   \vspace{-0.3em}
   \item 
   We develop a relaxed version of CROWN, linear bound propagation (LBP), which has better scalability. We demonstrate LBP can be used to obtain tighter bounds than IBP on both normally trained networks or IBP trained networks.
   \vspace{-0.3em}
   \item 
   We prove IBP is a special case of CROWN and LBP. The reason that CROWN gives looser bounds than IBP on IBP trained networks is that CROWN chooses bad bounding lines when making linear relaxations of the nonlinear activation functions. We prove CROWN and LBP are always tighter than IBP if they adopt the tight strategy to choose bounding lines as shown in Figure~\ref{fig:bounding_line_strategies}.
   \vspace{-1.5em}
   \item We propose to use a new activation function, parameterized ramp function (ParamRamp), to train verifiably robust networks. Compared with ReLU, where most neurons become dead during training, ParamRamp brings more diversity of neuron status. Our experiments demonstrate networks with ParamRamp activation achieve state-of-the-art verified $l_\infty$ robustness on MNIST, CIFAR-10 and Tiny-ImageNet.
\end{enumerate}
\vspace{-1.5em}
\section{Background and Related Work}
\vspace{-0.5em}
In this section, we start by giving definition of an $m$-layer feed-forward neural network 
and then briefly introduce the concept of robustness verification. 
Next we present interval bound propagation, which is used to train networks with best verified errors.
Finally we review two state-of-the-art verifiable adversarial training methods~\cite{gowal2019scalable, zhang2020towards} that are most related to our work.
\vspace{-1em}
\paragraph{Definition of an $m$-layer feed-forward network.}
\begin{align}
   \small
   \begin{split}
      {\vz}^{(k)} &= {\mW}^{(k)} {\va}^{(k-1)} + {\vb}^{(k)}, {\va}^{(k)} = \sigma({\vz}^{(k)}), \\
      k&=1,2,\cdots,m.
   \end{split}
   \vspace{-0.5em}
\end{align}
${\mW}^{(k)}, {\vb}^{(k)}, {\va}^{(k)}, {\vz}^{(k)}$ are the weight matrix, bias, activation and pre-activation of the $k$-th layer in the network, respectively. $\sigma$ is the elementwise activation function. Note that we always assume $\sigma$ is a monotonic increasing function in rest part of the paper. $\va^{(0)} = \vx$ and $\vz^{(m)}$ are the input and output of the network. We also use $n_k$ to denote the number of neurons in the $k$-th layer and $n_0$ is the dimension of the input. 
Although this network only contains fully connected layers, our discussions on this network in rest part of the paper readily generalize to convolutional layers as they are essentially a linear transformation as well~\cite{boopathy2019cnn}.

\vspace{-1.5em}
\paragraph{Robustness verification.}
Robustness verification aims to guarantee a neural network gives consistent predictions for all inputs in some set, typically an $l_p$ ball around the original input: $\mathbb{B}_p(\vx_0, \epsilon) = \{\vx \,|\, ||\vx-\vx_0||_p \leq \epsilon\}$, where $\vx_0$ is the clean image.
The key step is to compute the lower and upper bounds of the output logits $\vz^{(m)}$ (or the lower bound of the margin between ground truth class and other classes as defined in~\eqref{eqn:margin}) when the input can take any value in $\mathbb{B}_p(\vx_0, \epsilon)$. We can guarantee that the network gives correct predictions for all inputs in $\mathbb{B}_p(\vx_0, \epsilon)$ if the lower bound of the ground truth class is larger than the upper bounds of all the other classes (or the lower bound of the margin is greater than $0$). 
The verified robustness of a network is usually measured by the verified error: The percentage of images that we can not guarantee that the network always gives correct predictions for inputs in $\mathbb{B}_p(\vx_0, \epsilon)$.
Note that the verified error not only depends on the network and the allowed perturbation of the input, but also the method we use to compute bounds for the output. CROWN and IBP are the two bounding techniques that are most related to our work.
We briefly walk through CROWN in Section~\ref{sec:relaxed_crown} and introduce IBP right below.
\vspace{-1.5em}
\paragraph{Interval bound propagation.}
Assume we know the lower and upper bounds of the activation of the $(k-1)$-th layer: $\hat{\vl}^{(k-1)} \leq \va^{(k-1)} \leq \hat{\vu}^{(k-1)}$. Then IBP computes bounds of $\vz^{(k)}$, $\vl^{(k)}$ and $\vu^{(k)}$, in the following way:
\begin{align}
   \small
   \begin{split}
   \vl^{(k)}&=relu({\mW}^{(k)}) \hat{\vl}^{(k-1)} + neg({\mW}^{(k)}) \hat{\vu}^{(k-1)} + {\vb}^{(k)}, \\
   \vu^{(k)}&=relu({\mW}^{(k)}) \hat{\vu}^{(k-1)} + neg({\mW}^{(k)}) \hat{\vl}^{(k-1)} + {\vb}^{(k)},
   \end{split}
   \vspace{-1em}
\end{align}
where $relu$ is the elementwise ReLU function and $neg$ is the elementwise version of the function $neg(x) = x$, if $x\leq0$; $neg(x) = 0$, else.
Next, bounds of $\va^{(k)}$, $\hat{\vl}^{(k)}$ and $\hat{\vu}^{(k)}$, can be computed by
\begin{align}
   \small
   \begin{split}
   \hat{\vl}^{(k)} = \sigma({\vl}^{(k)}), \hat{\vu}^{(k)} = \sigma({\vu}^{(k)}).
   \end{split}
\end{align}
IBP repeats the above procedure from the first layer and computes bounds layer by layer until the final output as shown in Figure~\ref{fig:ibp_vs_lbp}. Bounds of $\va^{(0)}=\vx$ is known if the allowed perturbation is in an $l_\infty$ ball. Closed form bounds of $\vz^{(1)}$ can be computed using Holder's inequality as shown in~\eqref{eqn:closed_form_bound_z^k} if the allowed perturbation is in a general $l_p$ ball. 



\vspace{-1em} 
\paragraph{Train Verifiably Robust Networks.}
Verifiable adversarial training first use some robustness verification method to compute a lower bound $\vl^{\vomega}$ of the margin $\vomega$ between ground truth class $y$ and other classes:
\begin{align}
   \small
   \begin{split}
   &\vomega_i(\vx) = \vz^{(m)}_y(\vx) - \vz^{(m)}_i(\vx), i=1,2,\cdots,n_m. \label{eqn:margin}\\
   &\vl^{\vomega}(\vx_0, \epsilon) \leq \vomega(\vx), \forall \vx \in \mathbb{B}_p(\vx_0, \epsilon).
   \end{split}
\end{align}
Here we use ``$\leq$'' to denote elementwise less than or equal to. For simplicity, we won't differentiate operators between vectors and scalars in rest part of the paper when no ambiguity is caused.
Gowal \etal~\cite{gowal2019scalable} propose to use IBP to compute the lower bound $\vl_{\text{IBP}}^{\vomega}(\vx_0, \epsilon)$ and minimize the following loss during training:
\begin{align}
   \label{eqn:IBP_loss}
   \small
   \begin{split}
   \mathop{\mathbb{E}}_{(\vx_0,y)\in\mathcal{X}} \kappa L(\vz^{(m)}(\vx_0), y) + (1-\kappa) L(-\vl_{\text{IBP}}^{\vomega}(\vx_0, \epsilon), y),
   \end{split}
\end{align}
where $\mathcal{X}$ is the underlying data distribution, $\kappa$ is a hyper parameter to balance the two terms of the loss, and $L$ is the normal cross-entropy loss.
This loss encourages the network to maximize the margin between ground truth class and other classes.
Zhang \etal~\cite{zhang2020towards} argue that IBP bound is loose during the initial phase of training, which makes training unstable and hard to tune. They propose to use a convex combination of the IBP bound $\vl_{\text{IBP}}^{\vomega}$ and CROWN-IBP bound $\vl_{\text{C.-IBP}}^{\vomega}$ as the lower bound to provide supervision at the initial phase of training:
\begin{align}
   \label{eqn:convex_IBP_CROWN}
   \small
\begin{split}
   \vl^{\vomega} = (1-\beta) \vl_{\text{IBP}}^{\vomega} +\beta \vl_{\text{C.-IBP}}^{\vomega}.
\end{split}
\end{align}
The loss they use is the same as the one in \eqref{eqn:IBP_loss} except for replacing $\vl_{\text{IBP}}^{\vomega}$ with the new $\vl^{\vomega}$ defined in \eqref{eqn:convex_IBP_CROWN}. They design a schedule for $\beta$: It starts from $1$ and decreases to $0$ during training. Their approach achieves state-of-the-art verified errors on MNIST and CIFAR-10 datasets.
Xu \etal~\cite{xu2020automatic} propose a loss fusion technique to speed up the training process of CROWN-IBP and this enables them to train large networks on large datasets such as Tiny-ImageNet and Downscaled ImageNet.


\vspace{-0.5em}
\section{Relaxed CROWN}
\vspace{-0.5em}
\label{sec:relaxed_crown}
CROWN is considered an efficient robustness verification method compared with LP based methods \cite{weng2018towards, zhang2018crown, lyu2020fastened}, but these works only test CROWN on small multi-layer perceptrons with at most several thousand neurons in each hidden layer. Our experiment suggests that CROWN scales badly to large convolutional neural networks (CNNs): It consumes more than $12$ GB memory when verifying a single image from CIFAR-10 for a small $4$-layer CNN (See its detailed structure in 
{Appendix B.1}), which prevents it from utilizing modern GPUs to speed up computation. Therefore, it is crucial to improve CROWN's scalability to employ it on large networks. To this end, we develop a relaxed version of CROWN named \textbf{Linear Bound Propagation (LBP)}, whose computation complexity and memory cost grow linearly with the size of the network.
We first walk through the deduction process of the original CROWN. 

\vspace{-1em}
\paragraph{The original CROWN.}
Suppose we want to compute lower bound for the quantity $\mW^{obj} \vz^{(k)} + \vb^{obj}$. $\mW^{obj}$ and $\vb^{obj}$ are the weight and bias that connect $\vz^{(k)}$ to the quantity of interests. 
For example, the quantity becomes the margin $\vomega(\vx)$ if we choose appropriate $\mW^{obj}$ and set $\vb^{obj}=0, k=m$.
Assume we already know the bounds of pre-activation of the $(k-1)$-th layer:
\begin{align}
   \small
   \begin{split}
   \vl^{(k-1)} \leq \vz^{(k-1)} \leq \vu^{(k-1)}, \forall \vx \in \mathbb{B}_p(\vx_0, \epsilon).
   \label{eqn:lower_upper_bounds_of_z_k-1}
   \end{split}
\end{align} 
Next CROWN finds two linear functions of $\vz^{(k-1)}$ to bound $\va^{(k-1)}=\sigma(\vz^{(k-1)})$ in the intervals determined by $\vl^{(k-1)}, \vu^{(k-1)}$.
\begin{align}
   \small
   \begin{split}
   &\vh^{(k-1)L}(\vz^{(k-1)}) \leq \sigma(\vz^{(k-1)}) \leq \vh^{(k-1)U}(\vz^{(k-1)}), \\
   &\forall \, \vl^{(k-1)} \leq \vz^{(k-1)} 
   \leq \vu^{(k-1)},
   \end{split}
   \label{eqn:h_k-1}
\end{align}
where
\begin{align}
   \small
   \begin{split}
   \vh^{(k-1)L}(\vz^{(k-1)}) &= \vvs^{(k-1)L} * \vz^{(k-1)} + \vt^{(k-1)L},\\
   \vh^{(k-1)U}(\vz^{(k-1)}) &= \vvs^{(k-1)U} * \vz^{(k-1)} + \vt^{(k-1)U}.
   \end{split}
   \label{eqn:bounding_lines_of_k-1_layer}
\end{align}
Here we use ``$*$" to denote elementwise product.
$\vvs^{(k-1)L/U}, \vt^{(k-1)L/U}$ are constant vectors of the same dimension of $\vz^{(k-1)}$.
We use $L,U$ in the superscripts to denote quantities related to lower bounds and upper bounds, respectively.
We also use $L/U$ in the superscripts to denote ``lower bounds or upper bounds".
The linear functions $\vh^{(k-1)L/U}(\vz^{(k-1)})$ are also called bounding lines, as they bound the nonlinear function $\sigma(\vz^{(k-1)})$ in the intervals determined by $\vl^{(k-1)}, \vu^{(k-1)}$. 
See Figure~\ref{fig:bounding_line_strategies} for a visualization of different strategies to choose bounding lines. 
\begin{figure}[t]
   \centering
   \subfigure[CROWN vs Relaxed-CROWN-2]{
   \label{fig:crown_vs_relax}
   \includegraphics[width=0.99\columnwidth]{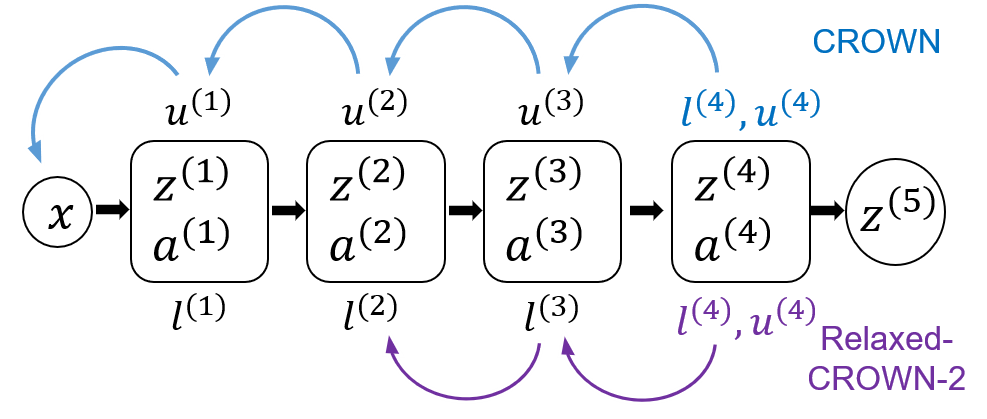}} \\
   \subfigure[IBP vs LBP]{
   \label{fig:ibp_vs_lbp}
   \includegraphics[width=0.99\columnwidth]{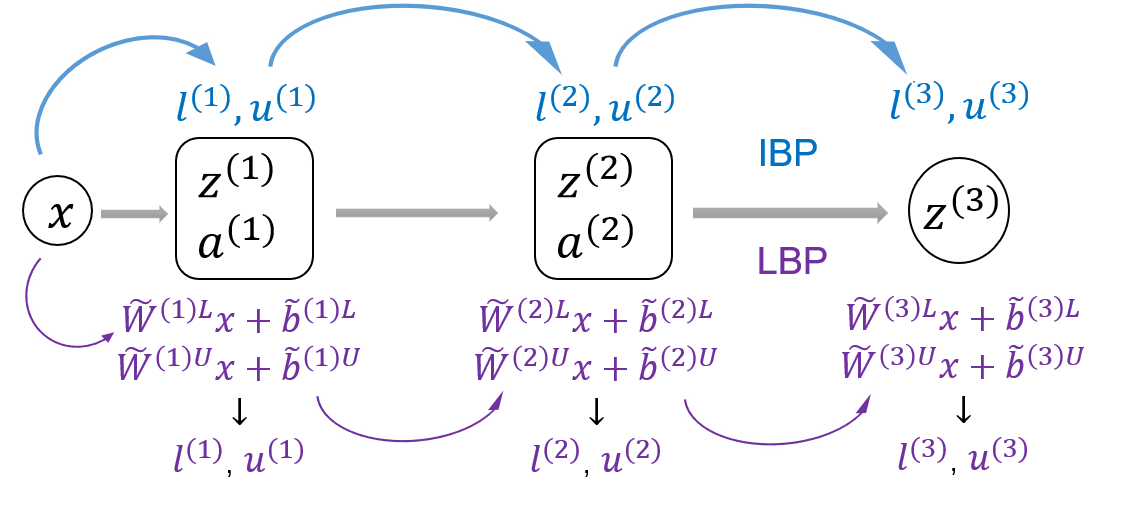}}
   \caption{Illustration of CROWN, Relaxed-CROWN, IBP and LBP. (a) shows how CROWN and Relaxed-CROWN-2 compute bounds for the $4$-th layer of a $5$ layer network. 
   (b) shows how IBP and LBP compute bounds layer by layer for a $3$ layer network.} 
   \vspace{-1em}
\end{figure}
Next CROWN utilizes these bounding lines to build a linear function of $\vz^{(k-1)}$ to lower bound $\mW^{obj} \vz^{(k)} + \vb^{obj}$:
\begin{align}
   \small
   \begin{split}
   \mW^{obj} \vz^{(k)} + \vb^{obj} \geq \mW^{(k,k-1)L} \vz^{(k-1)} + \vb^{(k,k-1)L}.
   \end{split}
\end{align}
See the detailed formulas of $\mW^{(k,k-1)L}, \vb^{(k,k-1)L}$ in 
{Appendix A.1}. 
In the same manner, CROWN builds a linear function of $\vz^{(k-2)}$ to lower bound $\mW^{(k,k-1)L} \vz^{(k-1)} + \vb^{(k,k-1)L}$ if bounds of $\vz^{(k-2)}$ are known.
CROWN repeats this procedure: It back-propagates layer by layer until the first layer $\vz^{(1)}$ as shown in Figure~\ref{fig:crown_vs_relax}:
\vspace{-0.25em}
\begin{align}
   \small
   \begin{split}
   &\mW^{obj} \vz^{(k)} + \vb^{obj} \geq \mW^{(k,k-1)L} \vz^{(k-1)} + \vb^{(k,k-1)L} \geq \cdots\\
   &\mW^{(k,k-2)L} \vz^{(k-2)} + \vb^{(k,k-2)L}
   \geq \mW^{(k,1)L} \vz^{(1)} + \vb^{(k,1)L}.
   \end{split}
   \label{eqn:z_1_to_z^k}
\end{align}
\vspace{-0.25em}
Notice $\vz^{(1)} = \mW^{(1)} \vx + \vb^{(1)}$. We plug it in the last term of \eqref{eqn:z_1_to_z^k} and obtain a linear function of $\vx$.
\begin{align}
   \small
\begin{split}
   \mW^{obj} \vz^{(k)} + \vb^{obj} &\geq \tilde{\mW}^{(k)L} \vx + \tilde{\vb}^{(k)L},
   \label{eqn:x_to_z^k}
\end{split}
\end{align}
where $\tilde{\mW}^{(k)L} = \mW^{(k,1)L} \mW^{(1)}, \tilde{\vb}^{(k)L} = \mW^{(k,1)L} \vb^{(1)} + \vb^{(k,1)L}$. Now we can compute the closed-form lower bound of $\mW^{obj} \vz^{(k)} + \vb^{obj}$ through Holder's inequality:
\begin{align}
   \small
   \begin{split}
   &\mW^{obj} \vz^{(k)} + \vb^{obj} 
   \geq \tilde{\mW}^{(k)L} \vx + \tilde{\vb}^{(k)L} \geq \\
   &\tilde{\mW}^{(k)L} \vx_0+\tilde{\vb}^{(k)L} - \epsilon ||\tilde{\mW}^{(k)L}||_q, \forall \, \vx \in \mathbb{B}_p(\vx_0, \epsilon),
   \end{split}
   \label{eqn:closed_form_bound_z^k}
\end{align}
where $1/p+1/q=1$ and $||\tilde{\mW}^{(k)L}||_q$ denotes a column vector that is composed of the $q$-norm of every row in $\tilde{\mW}^{(k)L}$.
We can compute a linear function of $\vx$ to upper bound $\mW^{obj} \vz^{(k)} + \vb^{obj}$ in the same manner and then compute its closed-form upper bound. See details in 
{Appendix A.1}.

Let's review the process of computing bounds for $\mW^{obj} \vz^{(k)} + \vb^{obj}$. It requires us to know the bounds of the previous $(k-1)$ layers: $\vz^{(1)}, \vz^{(2)}, \cdots, \vz^{(k-1)}$. We can fulfill this requirement by starting computing bounds from the first layer $\vz^{(1)}$, and then computing bounds layer by layer in a forward manner until the $(k-1)$-th layer. 
Therefore, 
the computation complexity of CROWN is of the order $\mathcal{O}(m^2)$.
And its memory cost is of the order $\mathcal{O}(\max_{k \neq v} n_{k} n_{v})$, where $0 \leq k,v \leq m$, and $n_k$ is the number of neurons in the $k$-th layer. 
This is because we need to record a weight matrix $\mW^{(k,v)}$ between any two layers as shown in \eqref{eqn:z_1_to_z^k}. This makes CROWN difficult to scale to large networks. To this end, we propose a relaxed version of CROWN in the next paragraph.
\vspace{-1em}
\paragraph{Relaxed CROWN.}
As the same in the above CROWN deduction process, suppose we want to compute bounds for the quantity $\mW^{obj} \vz^{(k)} + \vb^{obj}$. In the original CROWN process, we first compute linear functions of $\vx$ to bound the pre-activation of the first $(k-1)$ layers:
\begin{align}
   \small
   \begin{split}
   \tilde{\mW}^{(v)L} \vx + \tilde{\vb}^{(v)L} \leq \vz^{(v)} \leq \tilde{\mW}^{(v)U} \vx + \tilde{\vb}^{(v)U}, \\
   \forall \, \vx \in \mathbb{B}_p(\vx_0, \epsilon), v = 1,2,\cdots, k-1,
   \end{split}
   \label{eqn:linear_x_to_bound_z^v}
\end{align}
and use these linear functions of $\vx$ to compute closed-form bounds for the first $(k-1)$ layers.
We argue that in the back-propagation process in \eqref{eqn:z_1_to_z^k}, we don't need to back-propagate to the first layer. We can stop at any intermediate layer and plug in the linear functions in \eqref{eqn:linear_x_to_bound_z^v} of this intermediate layer to get a linear function of $\vx$ to bound $\mW^{obj} \vz^{(k)} + \vb^{obj}$. Specifically, assume we decide to back-propagate $v$ layers:
\begin{align}
   \small
   \begin{split}
   \mW^{obj} \vz^{(k)} + \vb^{obj} \geq \mW^{(k,k-1)L} \vz^{(k-1)} + \vb^{(k,k-1)L} \\
   \geq \cdots  \geq \mW^{(k,k-v)L} \vz^{(k-v)} + \vb^{(k,k-v)L}, v < k.
   \end{split}
   \label{eqn:z_k-v_to_z^k}
\end{align}
We already know 
\begin{align*}
   \small
\begin{split}
   \tilde{\mW}^{(k-v)L} \vx + \tilde{\vb}^{(k-v)L} \leq \vz^{(k-v)} \leq \tilde{\mW}^{(k-v)U} \vx + \tilde{\vb}^{(k-v)U}.
\end{split}
\end{align*}
We can directly plug it to \eqref{eqn:z_k-v_to_z^k} to obtain a lower bound of $\mW^{obj} \vz^{(k)} + \vb^{obj}$:
\begin{align}
   \small
\begin{split}
   &\mW^{obj} \vz^{(k)} + \vb^{obj} \geq \\
   &relu(\mW^{(k,k-v)L}) [\tilde{\mW}^{(k-v)L} \vx + \tilde{\vb}^{(k-v)L}] + \vb^{(k,k-v)L} \\
   & +neg(\mW^{(k,k-v)L})[\tilde{\mW}^{(k-v)U} \vx + \tilde{\vb}^{(k-v)U}].
   \label{eqn:relaxed_crown_v}
\end{split}
\end{align}
Now the last line of \eqref{eqn:relaxed_crown_v} is already a linear function of $\vx$ and we can compute the closed-form lower bound of $\mW^{obj} \vz^{(k)} + \vb^{obj}$ in the same manner as shown in \eqref{eqn:closed_form_bound_z^k}. The upper bound of $\mW^{obj} \vz^{(k)} + \vb^{obj}$ can also be computed by back-propagating only $v$ layers in the same gist. 

We have shown we can only back-propagate $v$ layers, instead of back-propagating to the first layer, when computing bounds for the $k$-th layer. In fact, we can only back-propagate $v$ layers when computing bounds for any layer. If the layer index $k$ is less than or equal to $v$, we just back-propagate to the first layer. In other words, we back-propagate at most $v$ layers when computing bounds for any layer in the process of CROWN. We call this relaxed version of CROWN  \textbf{Relaxed-CROWN-$v$}. 
See a comparison of CROWN and Relaxed-CROWN in Figure~\ref{fig:crown_vs_relax}.
\vspace{-1em}
\paragraph{Linear Bound Propagation.}
We are particularly interested in the special case of Relaxed-CROWN-$1$, namely, we only back-propagate $1$ layer in the process of CROWN. This leads us to the following theorem. 
\begin{theorem}
\label{thm:lbp}
Assume we already know two linear functions of $\vx$ to bound $\vz^{(k-1)}$:
\begin{align}
   \small
   \begin{split}
   \tilde{\mW}^{(k-1)L} \vx + \tilde{\vb}^{(k-1)L} \leq \vz^{(k-1)} \leq \tilde{\mW}^{(k-1)U} \vx + \tilde{\vb}^{(k-1)U}.
   \nonumber
\end{split}
\end{align}
We then compute the closed-form bounds $\vl^{(k-1)}, \vu^{(k-1)}$ of $\vz^{(k-1)}$ using these two linear functions, and choose two linear functions $\vh^{(k-1)L}(\vz^{(k-1)}), \vh^{(k-1)U}(\vz^{(k-1)})$ to bound $\va^{(k-1)}=\sigma(\vz^{(k-1)})$ as shown in \eqref{eqn:bounding_lines_of_k-1_layer}.
Then under the condition that
$\vvs^{(k-1)L} \geq 0, \vvs^{(k-1)U} \geq 0$,
$\vz^{(k)} = \mW^{(k)} \sigma(\vz^{(k-1)}) + \vb^{(k)}$ can be bounded by
\begin{align}
\small
\begin{split}
   \tilde{\mW}^{(k)L} \vx + \tilde{\vb}^{(k)L} \leq \vz^{(k)} \leq \tilde{\mW}^{(k)U} \vx + \tilde{\vb}^{(k)U},
\end{split} 
\end{align}
where
\begin{align}
   \small
   \begin{split}
   \tilde{\mW}^{(k)L} 
   &= [relu(\mW^{(k)}) * \vvs^{(k-1)L}] \tilde{\mW}^{(k-1)L} 
   \\&\;\;\;\;+ [neg(\mW^{(k)}) * \vvs^{(k-1)U}] \tilde{\mW}^{(k-1)U}, \\
   \tilde{\vb}^{(k)L} 
   &= \vb^{(k)} + [relu(\mW^{(k)}) * \vvs^{(k-1)L}] \tilde{\vb}^{(k-1)L} \\
   &\;\;\;\;+ [neg(\mW^{(k)}) * \vvs^{(k-1)U}] \tilde{\vb}^{(k-1)U} \\
   &\;\;\;\;+ relu(\mW^{(k)}) \vt^{(k-1)L} + neg(\mW^{(k)}) \vt^{(k-1)U},
   \end{split}
\end{align}
where the operator ``$*$'' between a matrix $\mW$ and a vector $\vvs$ is defined as $(\mW*\vvs)_{ij} = \mW_{ij} \vvs_j$. 
\end{theorem}
We refer readers to 
{Appendix A.3} for the formulas of $\tilde{\mW}^{(k)U}, \tilde{\vb}^{(k)U}$ and the proof of Theorem~\ref{thm:lbp}. 
Note that 
the condition $\vvs^{(k-1)L} \geq 0, \vvs^{(k-1)U} \geq 0$
in Theorem~\ref{thm:lbp} is not necessary. We impose this condition because it simplifies the expressions of $\tilde{\mW}^{(k)L}, \tilde{\vb}^{(k)L},$ and it generally holds true when people choose bounding lines.

The significance of Theorem~\ref{thm:lbp} is that it allows us to compute bounds starting from the first layer $\vz^{(1)}$, which can be bounded by $\mW^{(1)} \vx + \vb^{(1)} \leq \vz^{(1)} \leq \mW^{(1)} \vx + \vb^{(1)}$, and then compute bounds layer by layer in a forward manner until the final output just like IBP. 
The computation complexity is reduced to $\mathcal{O}(m)$ and memory cost is reduced to $\mathcal{O}(n_0 \max\{n_1, n_2, \cdots, n_m\})$,
since we only need to record a matrix $\tilde{\mW}^{(k)}$ from the input $\vx$ to every intermediate layer $\vz^{(k)}$. We call this method \textbf{Linear Bound Propagation (LBP)}, which is equivalent to Relaxed-CROWN-$1$.
See a comparison of LBP and IBP in Figure~\ref{fig:ibp_vs_lbp}.
As expected, there is no free lunch. As we will show in the next section, the reduction of computation and memory cost of LBP makes it less tight than CROWN. 
Although developed from a different perspective, we find LBP similar to the forward mode in the work~\cite{xu2020automatic}. See a detailed comparison between them in Appendix A.3.

Zhang~\etal \cite{zhang2020towards} propose to compute bounds for the first $(m-1)$ layers using IBP and then use CROWN to compute bounds for the last layer to obtain tighter bounds of the last layer. The resulting method is named CROWN-IBP. In the same gist, we can use LBP to compute bounds for the first $(m-1)$ layers and then use CROWN to compute bounds for the last layer. We call this method \textbf{CROWN-LBP}. 
\section{Relationship of IBP, LBP and CROWN}
\vspace{-0.5em}
\label{sec:relationship_ibp_lbp_crown}
In Section~\ref{sec:relaxed_crown}, we develop a relaxed version of CROWN, LBP. 
In this section, we study the relationship between IBP, LBP and CROWN, and investigate why CROWN gives looser bounds than IBP on IBP trained networks~\cite{zhang2020towards}.

First, we manage to prove IBP is a special case of CROWN and LBP where the bounding lines are chosen as constants as shown in Figure~\ref{fig:constant_bdl}:
\begin{align}
\small
\begin{split}
   \vh^{(k)L}(\vz^{(k)}) = \sigma(\vl^{(k)})&,
   \vh^{(k)U}(\vz^{(k)}) = \sigma(\vu^{(k)}),\\
   k=1,2,&\cdots,m-1.
\end{split}
\label{eqn:IBP_bounding_lines}
\end{align}
In other words, CROWN and LBP degenerate to IBP when they choose constant bounding lines for every neuron in every layer. See the proof of this conclusion in 
{Appendix A.5}.
On the other hand, Lyu \etal~\cite{lyu2020fastened} prove tighter bounding lines lead to tighter bounds in the process of CROWN, where $\tilde{\vh}_i^{(k)L/U}(\vz_i^{(k)})$ is defined to be tighter than $\hat{\vh}_i^{(k)L/U}(\vz_i^{(k)})$ in the interval $[\vl^{(k)}_i, \vu^{(k)}_i]$ if
\begin{align}
   \small
   \begin{split}
   \hat{\vh}_i^{(k)L}(\vz_i^{(k)}) \leq \tilde{\vh}_i^{(k)L}(\vz_i^{(k)}), &\tilde{\vh}_i^{(k)U}(\vz_i^{(k)}) \leq \hat{\vh}_i^{(k)U}(\vz_i^{(k)}), \\
   \forall \vz_i^{(k)} \in& [\vl^{(k)}_i, \vu^{(k)}_i].
   \end{split}
\end{align}
We manage to prove it is also true for LBP in 
{Appendix A.3}. Therefore, if CROWN and LBP adopt the 
tight strategy in Figure~\ref{fig:zero_bdl} to choose bounding lines, which is guaranteed to be tighter than the constant bounding lines 
in a specified interval, CROWN and LBP are guaranteed to give tighter bounds than IBP. 
We formalize this conclusion and include conclusions for CROWN-IBP and CROWN-LBP in the following theorem.
\begin{figure}[tb]
   \centering
   \includegraphics[width=0.99\columnwidth]{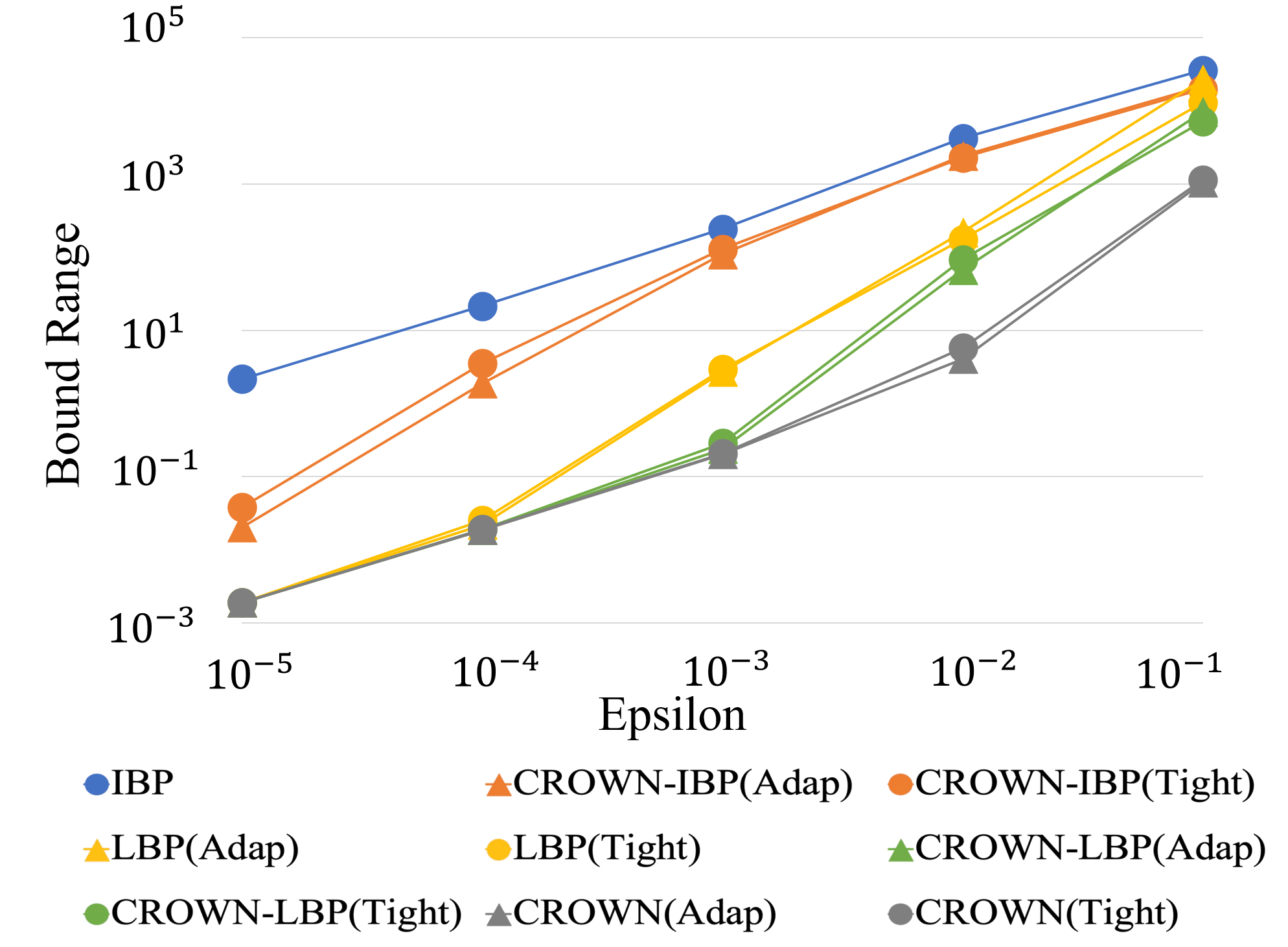}
   \caption{Tightness comparison of IBP, CROWN-IBP, LBP, CROWN-LBP, CROWN on a normally trained MNIST classifier. ``Bound Range'' is mean of 
   $\vu^{(m)}-\vl^{(m)}$. The mean is taken over the $10$ output logits and averaged over $100$ test images in MNIST. ``Epsilon'' is the radius of the $l_{\infty}$ ball. ``Adap'' and ``Tight'' are the adaptive and tight strategies as shown in Figure~\ref{fig:bounding_line_strategies}. 
   }
   \label{fig:compare_ibp_lbp_crown_on_normal_model}
   \vspace{-1em}
\end{figure}
\vspace{-0.5em}
\begin{theorem}
   \label{thm:compare_5_methods}
   Assume the closed-form bounds of the last layer computed by IBP, CROWN-IBP, LBP, CROWN-LBP, and CROWN are $\vl^{(m)}_I$, $\vu^{(m)}_I$; $\vl^{(m)}_{CI}$, $\vu^{(m)}_{CI}$; $\vl^{(m)}_L$, $\vu^{(m)}_L$; $\vl^{(m)}_{CL}$, $\vu^{(m)}_{CL}$; $\vl^{(m)}_C$, $\vu^{(m)}_C$, respectively. And CROWN-IBP, LBP, CROWN-LBP, CROWN adopt the 
   tight strategy to choose bounding lines as shown in Figure~\ref{fig:zero_bdl}. Then we have
   \begin{align}
      \small
      \begin{split}
      \vl^{(m)}_I &\leq \{\vl^{(m)}_{L} , \vl^{(m)}_{CI}\} \leq \vl^{(m)}_{CL} \leq \vl^{(m)}_C, \\
      \vu^{(m)}_I &\geq \{\vu^{(m)}_{L} , \vu^{(m)}_{CI}\} \geq \vu^{(m)}_{CL} \geq \vu^{(m)}_C,
      \end{split}
   \end{align} 
   where the sets in the inequalities mean that the inequalities hold true for any element in the sets.
\end{theorem}
See proof of Theorem~\ref{thm:compare_5_methods} in 
{Appendix A.6}.
Now we can answer the question proposed at the beginning of this section. 
The reason that CROWN gives looser bounds than IBP \cite{zhang2020towards} is because CROWN uses the 
adaptive strategy as shown in Figure~\ref{fig:adaptive_bdl_1} and~\ref{fig:adaptive_bdl_2} to choose bounding lines by default.
The lower bounding line chosen in the adaptive strategy for an unstable neuron is not always tighter than the one chosen by the constant strategy adopted by IBP.
Zhang \etal~\cite{zhang2018crown} emperically show the 
adaptive strategy gives tighter bounds for normally trained networks. 
An intuitive explanation is that this strategy minimizes the area between the lower and upper bounding lines in the interval, but there is no guarantee for this intuition. 
On the other hand, for IBP trained networks, the loss is optimized at the point where bounding lines are chosen as constants. Therefore we should choose the same constant bounding lines or tighter bounding lines for LBP or CROWN when verifying IBP trained networks, which is exactly what we are doing in the 
tight strategy.
\begin{table}[tb]
   \centering
   \caption{Mean lower bound of the margin defined in~\eqref{eqn:margin} and verified errors obtained by IBP, CROWN-IBP(C.-IBP), LBP, CROWN-LBP(C.-LBP) on an IBP trained CIFAR-10 classifier. The network is trained with $\epsilon=8.8/255$ and tested with $\epsilon=8/255$ ($l_{\infty}$ norm).
   Results are taken over $100$ test images.
   }
   \label{tbl:compare_ibp_lbp_crown_on_robust_model}
   \resizebox{1\columnwidth}{!}{
   \begin{tabular}{c|c|c|c|c}
      \noalign{\hrule height 0.75pt}
      Adaptive & IBP & C.-IBP & LBP & C.-LBP \\
      Verified Err(\%) & 70.10 & 85.66 & 100 & 99.99\\
      Lower Bound & 2.1252 & -12.016 & -2.4586E5 & -1.5163E5\\
      \hline
      Tight & IBP & C.-IBP & LBP & C.-LBP \\
      Verified Err(\%) & 70.10 & 70.01 & 70.05 & 69.98\\
      Lower Bound & 2.1252 & 2.1520 & 2.1278 & 2.1521\\
      \noalign{\hrule height 0.75pt}
   \end{tabular}
   }
   \vspace{-1em}
\end{table}

We conduct experiments to verify our theory. 
We first compare IBP, LBP and CROWN on a normally trained MNIST classifier (See its detailed structures in 
{Appendix B.1}). 
Result is shown in Figure~\ref{fig:compare_ibp_lbp_crown_on_normal_model}. The average verification time for a single image of IBP, CROWN-IBP, LBP, CROWN-LBP, CROWN are 0.006s, 0.011s, 0.027s, 0.032s, 0.25s, respectively, tested on one NVIDIA GeForce GTX TITAN X GPU. 
We can see LBP is tighter than IBP while being faster than CROWN. 
And the adaptive strategy usually obtains tighter bounds than the tight strategy. 
See more comparisons of these methods in 
{Appendix B.2}.

Next, we compare them on an IBP trained network.
The network we use is called DM-large (See its detailed structure in 
{Appendix B.1}), which is the same model in the work\cite{zhang2020towards, gowal2019scalable}.
Results are shown in Table~\ref{tbl:compare_ibp_lbp_crown_on_robust_model}.
We don't test CROWN on this network because it exceeds GPU memory (12 GB) and takes about half an hour to verify a single image on one Intel Xeon E5-2650 v4 CPU.
We can see CROWN-IBP, LBP and CROWN-LBP give worse verified errors than IBP when adopting adaptive strategy to choose bounding lines, but give better results when adopting the tight strategy as guaranteed by Theorem~\ref{thm:compare_5_methods}.
However, we can see the improvement of LBP and CROWN-LBP over IBP and CROWN-IBP is small compared with the normally trained network. We investigate this phenomenon in the next section. 
\section{Parameterized Ramp Activation}
\begin{figure}[tb]
   \centering
   \includegraphics[width=1\columnwidth]{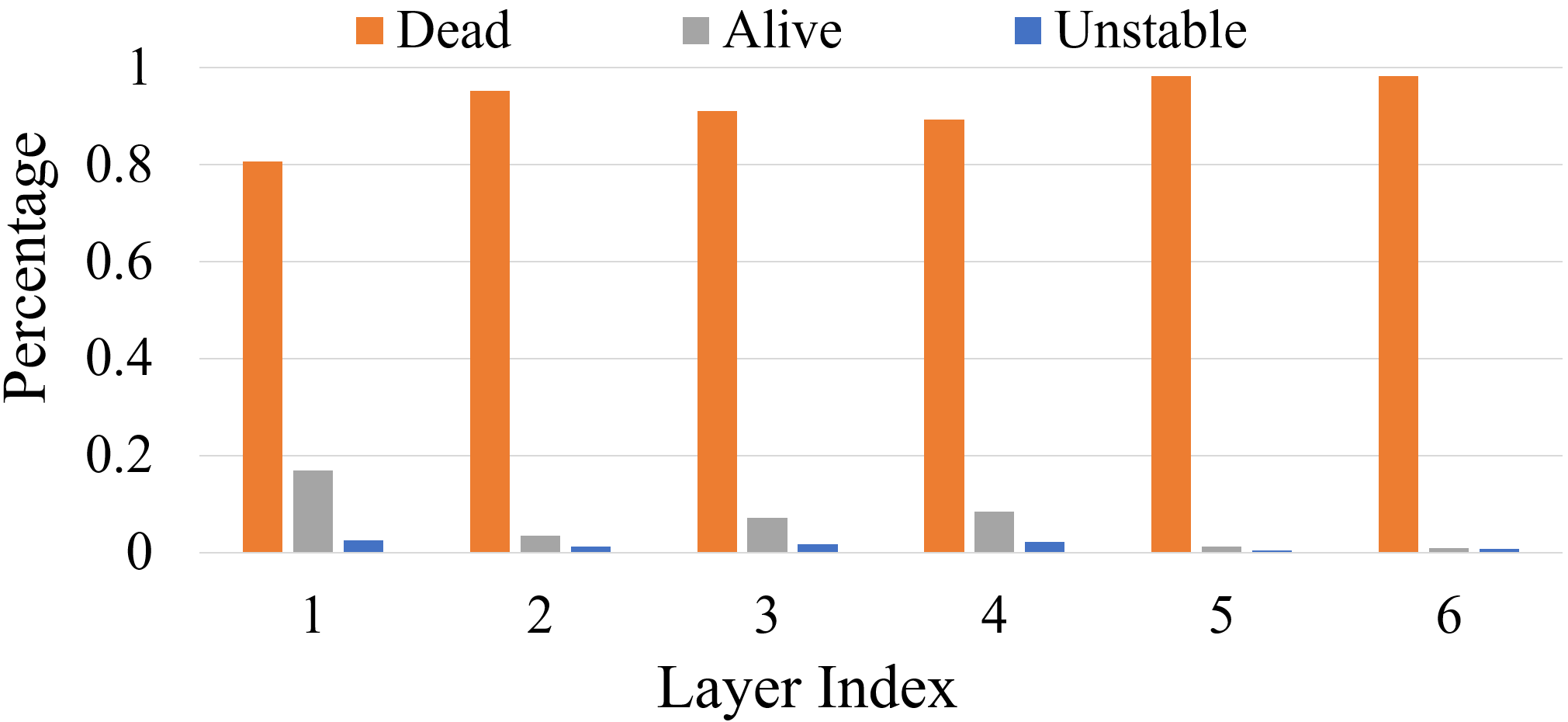}
   \caption{Neuron status of an IBP trained DM-large network at $\epsilon=2.2/255$ ($l_\infty $ norm) on CIFAR-10.
   Bounds are computed using IBP at $\epsilon=2/255$. The horizontal axis is the layer index, and the vertical axis is the percentage of every neuron status in the layer. The percentage is averaged over $100$ CIFAR-10 test images.}
   \label{fig:neuron_status_comparison}
   \vspace{-1em}
\end{figure}

\vspace{-0.5em}
This section starts by investigating the phenomenon discovered in Section~\ref{sec:relationship_ibp_lbp_crown}: Why the improvement of LBP and CROWN-LBP over IBP and CROWN-IBP is so small on the IBP trained network compared with the normally trained network.
Study of this phenomenon inspires us to design a new activation function to achieve 
lower verified errors.
\vspace{-1em}
\paragraph{Investigate the limited improvement of LBP.}
We argue that the limited improvement of LBP and CROWN-LBP is because most neurons are dead in IBP trained networks. Recall that we define three status of a ReLU neuron according to the range of its input in Figure~\ref{fig:bounding_line_strategies}: Dead, Alive, Unstable. 
We demonstrate neuron status in each layer of an IBP trained network in Figure~\ref{fig:neuron_status_comparison}.
We can see most neurons are dead. However, we find most neurons (more than 95\%) are unstable in a normally trained network.
For unstable neurons, bounding lines in the tight strategy adopted by LBP and CROWN are tighter than the constant bounding lines chosen by IBP. This explains why LBP and CROWN are several orders tighter than IBP for a normally trained network.
However, for dead neurons, the bounding lines chosen by LBP and CROWN are the same as those chosen by IBP, which explains the limited improvement of LBP and CROWN-LBP on IBP trained networks. We conduct experiments in 
{Appendix B.3} to further verify this explanation.

\begin{figure}[tb]
   \centering
   \includegraphics[width=0.9\columnwidth]{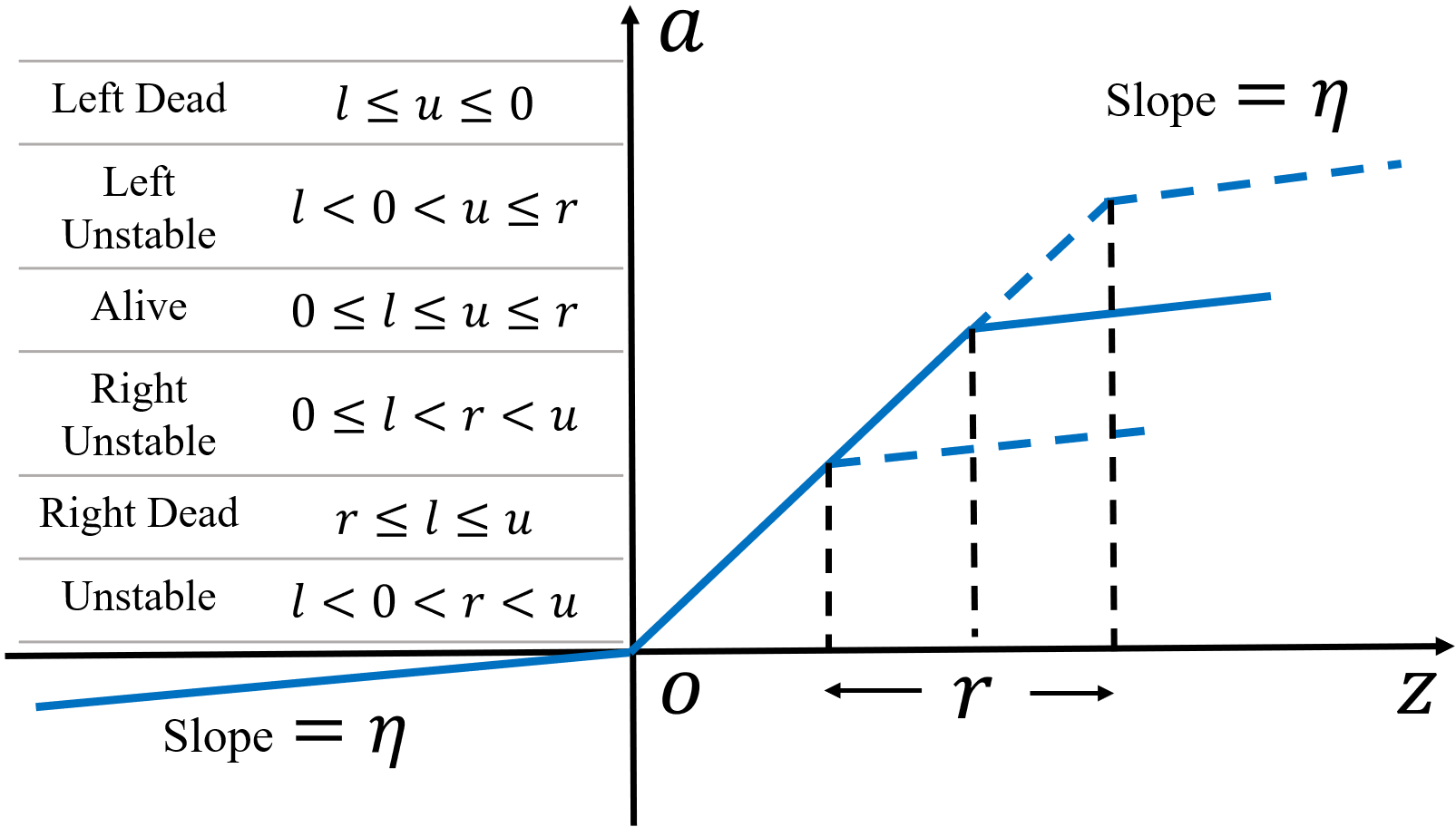}
   \caption{Parameterized Ramp (ParamRamp) function. The bending point $r$ is tunable. We can define six status of a neuron according to the input range $[l,u]$ as shown in the left side. See 
   {Appendix A.7} for how to choose bounding lines for ParamRamp.}
   \label{fig:param_ramp}
   \vspace{-1em}
\end{figure}

It seems reasonable that most neurons are dead in IBP trained networks, since dead neurons can block  perturbations from the input, which makes the network more robust. 
However, we argue that there are two major drawbacks caused by this phenomenon: 
First, gradients from both the normal cross-entropy loss and IBP bound loss in~\eqref{eqn:IBP_loss} can not back-propagate through dead neurons. This may prevent the network from learning at some point of the training process. 
Second, it restricts the representation capability of the network, since most activations are $0$ in intermediate layers.
\vspace{-2em}
\paragraph{Parameterized Ramp function.}
To mitigate these two problems, one simple idea is to use LeakyReLU instead of ReLU during training. We will consider this approach as the baseline and compare with it.
We propose to use a \textbf{Parameterized Ramp (ParamRamp)} function to achieve better result. The Parameterized Ramp function can be seen as a LeakyReLU function with the right part being bent flat at some point $r$, as shown in Figure~\ref{fig:param_ramp}. The parameter $r$ is tunable for every neuron. We include it to the parameters of the network and optimize over it during training. The intuition behind this activation function is that it provides another robust (function value changes very slowly with respect to the input) region on its right part. 
This right part has function values greater than $0$ and tunable, in comparison to the left robust region with function values close to $0$. 
Therefore during the IBP training process, a neuron has two options to become robust: to become either left dead or right dead as shown in Figure~\ref{fig:param_ramp}. 
This could increase the representation capability of the network while allow it to become robust. 
We compare effects of ReLU, LeakyReLU and ParamRamp functions in terms of training verifiably robust networks in the next section.
\section{Experiments}
\begin{table*}[t]
   \centering
   \vspace{-1.75em}
   \caption{\small{Errors of IBP trained and CROWN-IBP trained networks with different activations on CIFAR-10. We report errors on clean images(Clean: Percentage of images wrongly classified), IBP verified errors(IBP), CROWN-LBP verified errors(C.-LBP), and PGD attack errors(PGD: Percentage of images
   successfully attacked). 
   Experiments are conducted on $3$ variants of ParamRamp: Ramp(0), ParamRamp with $\eta=0$; Ramp(0.01), $\eta=0.01$; Ramp(0.01$\rightarrow$0), $\eta$ starts from $0.01$ and gradually decreases to $0$ during training.
   $3$ variants of ReLU are similarly designed,
   \eg, ReLU(0.01) means LeakyReLU with leakage slope 0.01. 
   Networks are trained at $\epsilon=2.2/255, 8.8/255$ and evaluated at $\epsilon=2/255, 8/255$ respectively. 
   Results of ReLU(0) are directly copied from the original works~\cite{gowal2019scalable, zhang2020towards}. We compute C.-LBP verified errors based on our re-run networks for these experiments. Therefore, C.-LBP verified errors are not comparable to IBP verified errors on these networks. 
   We also report results run on large ReLU networks in the work~\cite{xu2020automatic} at the right side.}}
   \label{tbl:cifar}
   \resizebox{2.09\columnwidth}{!}{  
   \begin{tabular}{c|l||c|c|c|c||c|c|c|c||c|c|c|c}
      \noalign{\hrule height 0.9pt}
      \multirow{2}{*}{\makecell{Training\\Method}} & \multirow{2}{*}{Activation} & \multicolumn{4}{c||}{Errors (\%) for $\epsilon=2/255$} & \multicolumn{4}{c||}{Errors (\%) for $\epsilon=8/255$}  & \multicolumn{4}{c}{Errors (\%) for $\epsilon=8/255$} \\ 
      \cline{3-14}
      &  & Clean & IBP & C.-LBP & PGD & Clean & IBP & C.-LBP & PGD & Model & Clean & IBP & PGD \\ 
      \noalign{\hrule height 0.8pt}
      \multirow{6}{*}{IBP} & ReLU(0) & 39.22 & 55.19 & 54.38 & 50.40 & 58.43 & 70.81 & 69.98 & 68.73 & \multirow{6}{*}{\makecell{CNN-7+BN\\Densenet\\WideResNet\\ResNeXt}} & \multirow{6}{*}{\makecell{57.95\\57.21\\58.07\\\textbf{56.32}}} & \multirow{6}{*}{\makecell{\textbf{69.56}\\69.59\\70.04\\70.41}} & \multirow{6}{*}{\makecell{\textbf{67.10}\\67.75\\67.23\\67.55}} \\ 
      & ReLU(0.01)  & \textbf{32.3} & 52.02 & 47.26 & 44.22 & 55.16 & 69.05 & 68.45 & 66.05 &  &  &  &  \\ 
      & ReLU(0.01$\rightarrow$0)& 34.6 & 53.77 & 51.62 & 46.71 & 55.62 & 68.32 & 68.22 & 65.29 &  &  &  &  \\ 
      \cline{2-10}
      & Ramp(0) & 36.47 & 53.09 & 52.28 & 46.52 & 56.32 & 68.89 & 68.82 & 63.89 &  &  &  &  \\ 
      & Ramp(0.01) & 33.45 & 48.39 & \textbf{47.19} & 43.87 & \textbf{54.16} & 68.26 & 67.78 & 65.06 &  &  &  &  \\ 
      & Ramp(0.01$\rightarrow$0) & 34.17 & \textbf{47.84} & 47.46 & \textbf{42.74} & 55.28 & \textbf{67.26} & \textbf{67.09} & \textbf{60.39} &  &  &  &  \\ 
      \noalign{\hrule height 0.8pt}
      \multirow{6}{*}{\makecell{CROWN\\-IBP}} & ReLU(0) & 28.48 & 46.03 & 45.04 & 40.28 & 54.02 & 66.94 & 66.69 & 65.42 & \multirow{6}{*}{\makecell{CNN-7+BN\\Densenet\\WideResNet\\ResNeXt}} & \multirow{6}{*}{\makecell{\textbf{53.71}\\56.03\\53.89\\53.85}} & \multirow{6}{*}{\makecell{\textbf{66.62}\\67.57\\67.77\\68.25}} & \multirow{6}{*}{\makecell{64.31\\65.09\\64.42\\\textbf{64.16}}} \\ 
      & ReLU(0.01) & 28.49 & 46.68 & 44.09 & 39.29 & 55.18 & 68.54 & 68.13 & 66.41 &  &  &  &  \\
      & ReLU(0.01$\rightarrow$0)& \textbf{28.07} & 46.82 & 44.40 & 39.29 & 63.88 & 72.28 & 72.13 & 70.34 &  &  &  &  \\
      \cline{2-10}
      & Ramp(0) & 28.48 & \textbf{45.67} & 44.03 & 39.43 & 52.52 & 65.24 & 65.12 & 62.51 &  &  &  &  \\ 
      & Ramp(0.01) & 28.63 & 46.17 & 44.28 & 39.61 & 52.15 & 66.04 & 65.75 & 63.85 &  &  &  &  \\ 
      & Ramp(0.01$\rightarrow$0)& 28.18 & 45.74 & \textbf{43.37} & \textbf{39.17} & \textbf{51.94} & \textbf{65.19} & \textbf{65.08} & \textbf{62.05} &  &  &  &  \\ 
      \noalign{\hrule height 0.9pt}
   \end{tabular}
   }
   \vspace{-0.5em}
\end{table*}
\begin{table}[t]
   \centering
   \caption{Comparison of ParamRamp and ReLU on MNIST dataset. Notations are the same as those in Table~\ref{tbl:cifar}. The networks are both trained and tested at $\epsilon=0.4$. See more experiments tested with different activation functions and at different $\epsilon$ in 
   {Appendix B.5}.}
   \label{tbl:mnist}
   \resizebox{1\columnwidth}{!}{  
   \begin{tabular}{c|l||c|c|c|c}
      \noalign{\hrule height 0.9pt}
      \multirow{2}{*}{\makecell{Training\\Method}} & \multirow{2}{*}{Activation} & \multicolumn{4}{c}{Errors (\%) for $\epsilon=0.4$} \\ 
      \cline{3-6}
      &  & Clean & IBP & C.-LBP & PGD \\ 
      \hline
      \multirow{2}{*}{IBP} & ReLU(0) & 2.74 & 14.80 & 16.13 & 11.14 \\ 
      & Ramp(0.01$\rightarrow$0) & 2.16 & {10.90} & {10.88} & 6.59 \\ 
      \hline
      \multirow{2}{*}{\makecell{CROWN\\-IBP}} & ReLU(0) & 2.17 & 12.06 & 11.90 & 9.47 \\ 
      & Ramp(0.01$\rightarrow$0) & 2.36 & {10.68} & {10.61} & 6.61\\
      \noalign{\hrule height 0.9pt}
   \end{tabular}
   }
   \vspace{-2em}
\end{table}
In this section, we conduct experiments to train verifiably robust networks using our proposed activation function, ParamRamp, and compare it with ReLU and LeakyReLU.
We use the loss defined in~\eqref{eqn:IBP_loss} and consider $l_{\infty}$ robustness in all experiments.
The experiments are conducted on $3$ datasets: MNIST, CIFAR-10, and Tiny-ImageNet.
For MNIST and CIFAR-10 datasets, we use the same DM-large network, and follow the same IBP training and CROWN-IBP training procedures in the works~\cite{gowal2019scalable, zhang2020towards}.
For the Tiny-ImageNet dataset, we follow the training procedure in the work~\cite{xu2020automatic}. The networks we train on Tiny-ImageNet are a $7$-layer CNN with Batch Normalization layers (CNN-7+BN) and a WideResNet.  
We refer readers to the original works
or 
{Appendix B.4} for detailed experimental set-ups and network structures.
During the training of ParamRamp networks, it is important to initialize the tunable parameters $r$ appropriately. We also find ParamRamp networks have overfitting problems in some cases. See how we initialize $r$ and solve the overfitting problem in Appendix B.4.
After training, we use IBP and CROWN-LBP with the tight strategy to compute verified errors. IBP verified errors allow us to compare results with previous works, and CROWN-LBP gives us the best verified errors as guaranteed in Theorem~\ref{thm:compare_5_methods}. 
CROWN is not considered because it exceeds GPU memory (12 GB) to verify a single image on the networks we use and is extremely slow running on CPU. We also use 200-step PGD attacks~\cite{madry2018towards} with 10 random starts to emperically evaluate robustness of the networks.

Results on CIFAR-10 and MNIST datasets are presented in Table~\ref{tbl:cifar} and Table~\ref{tbl:mnist}, respectively. 
We can see networks with ParamRamp activation achieve better verified errors, clean errors, and PGD attack errors than ReLU networks in almost all settings. 
And our proposed bound computation method, CROWN-LBP, can always provide lower verified errors than IBP. 
See more experiments for networks of different structures in 
{Appendix B.5}.
For Tiny-ImageNet dataset, the CNN-7+BN and WideResNet networks with ParamRamp activation achieve $84.99\%$ and $82.94\%$ IBP verified errors at $\epsilon=1/255$, respectively. To the best of our knowledge, $82.94\%$ is the best verified error at $\epsilon=1/255$ ever achieved on Tiny-ImageNet. See a comparison with ReLU networks from the work~\cite{xu2020automatic} in 
{Appendix B.5}.

\begin{figure}[t]
   \centering
   \includegraphics[width=1\columnwidth]{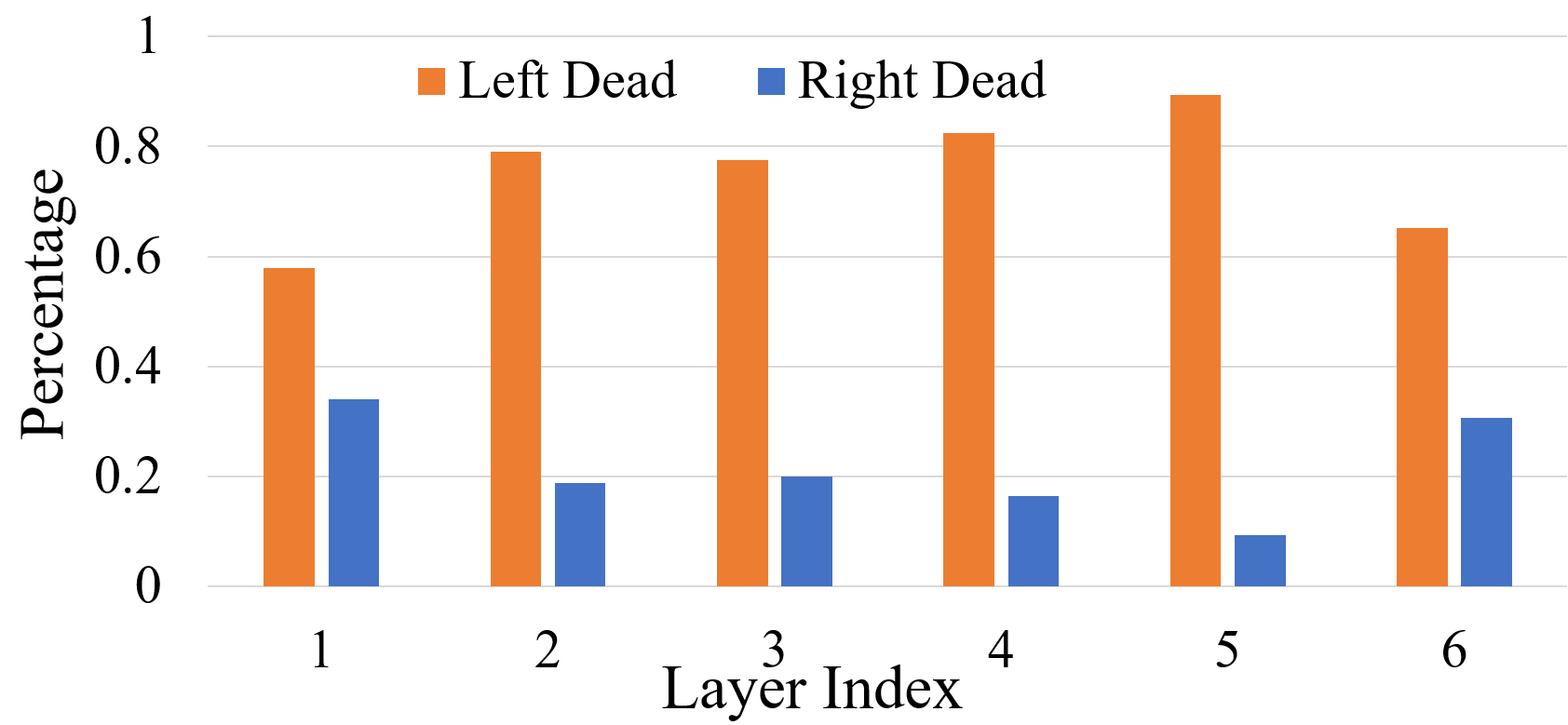}
   \caption{Neuron status of an IBP trained network on CIFAR-10 with ParamRamp activation. We only present left dead status and right dead status since most neurons are in these two status. The network is trained at $\epsilon=2.2/255$ ($l_\infty$ norm) and bounds are computed using IBP at $\epsilon=2/255$.}
   \label{fig:neuron_status_of_param_ramp}
   \vspace{-1.5em}
\end{figure}

ParamRamp activation brings additional parameters to the network. 
We are concerned about its computational overhead compared with ReLU networks. On MNIST, we find the average training time per epoch of a ParamRamp network is $1.09$ times of that of a ReLU network in IBP training, and is $1.51$ times in CROWN-IBP training.
We observe an overhead of similar level on CIFAR-10 and Tiny-ImageNet datasets. See a full comparison in 
{Appendix B.5}.   
Comparing ParamRamp with ReLU on the same network may not be convincing enough to demonstrate the superiority of ParamRamp, as it has additional parameters. 
We compare it with larger size ReLU networks trained in the work~\cite{xu2020automatic}. 
We report their results on CNN-7+BN, Densenet~\cite{huang2017densely}, WideResNet~\cite{zagoruyko2016wide} and ResNeXt~\cite{xie2017aggregated} in the right part of Table~\ref{tbl:cifar}. 
Despite being larger than the DM-large network with ParamRamp activation, these ReLU networks still can not obtain lower IBP verified errors than our model.
We think this is because ParamRamp activation brings more diversity of neuron status, which increases the representation capability of the network.
Recall that most neurons are dead in IBP trained ReLU networks as shown in Figure~\ref{fig:neuron_status_comparison}. 
We present neuron status of an IBP trained ParamRamp network in Figure~\ref{fig:neuron_status_of_param_ramp}. 
We can see although lots of neurons are still left dead, there is a considerable amount of neurons are right dead. 
Note that the activation value of right dead neurons are not $0$ and tunable. This allows the network to become robust while preserving representation capability.
See more neuron status comparisons of ReLU and ParamRamp networks in Appendix B.5.
\vspace{-0.5em}
\section{Conclusion}
\vspace{-0.5em}
\label{sec:conclusion}
We propose a new verification method, LBP, which has better scalability than CROWN while being tighter than IBP.
We further prove CROWN and LBP are always tighter than IBP when choosing appropriate bounding lines, and can be used to verify IBP trained networks to obtain lower verified errors. 
We also propose a new activation function, ParamRamp, to mitigate the problem that most neurons become dead in ReLU networks during IBP training.
Extensive experiments demonstrate networks with ParamRamp activation outperforms ReLU networks and achieve state-of-the-art $l_\infty$ verified robustness on MNIST, CIFAR-10 and Tiny-ImageNet datasets.
\vspace{-1em}
\paragraph{Acknowledgement.}
This work is partially supported by General Research Fund (GRF) of Hong Kong (No. 14203518), Collaborative Research Grant from SenseTime Group (CUHK Agreement No. TS1712093, and No. TS1711490), and the Shanghai Committee of Science and Technology, China (Grant No. 20DZ1100800). The authors thank useful discussions with Xudong Xu.



{\small
\bibliographystyle{ieee_fullname}
\bibliography{egbib}
}

\end{document}